\pdfoutput=1

\documentclass[11pt]{article}

\usepackage[preprint]{acl}

\usepackage{times}
\usepackage{latexsym}

\usepackage[T1]{fontenc}

\usepackage[utf8]{inputenc}

\usepackage{microtype}

\usepackage{inconsolata}

\usepackage{graphicx}

\usepackage{booktabs}       
\usepackage[ruled,vlined]{algorithm2e}
\usepackage{enumitem}       
\usepackage{amssymb}        
\usepackage{amsmath}  
\usepackage{amssymb}  
\usepackage{algorithm2e} 
\usepackage[hang,flushmargin]{footmisc}  

\title{Emergence: Overcoming Privileged Information Bias in Asymmetric Embodied Agents via Active Querying}

\author{Shaun Baek \and Sam Liu \and Joseph Ukpong \\
        Emory University}

\begin{document}
\maketitle

\begin{abstract}
Large Language Models (LLMs) act as powerful reasoning engines but struggle with ``symbol grounding'' in embodied environments, particularly when information is asymmetrically distributed. We investigate the \textit{Privileged Information Bias} (or ``Curse of Knowledge''), where a knowledgeable ``Leader'' agent fails to guide a sensor-limited ``Follower'' due to a lack of Theory of Mind. To quantify this phenomenon, we propose a novel \textbf{Asymmetric Assistive Reasoning framework} within AI2-THOR. Our experiments reveal a significant ``Success Gap'': while the Leader successfully perceives the target in 35.0\% of episodes, the collaborative team succeeds only 17.0\% of the time, implying that nearly 50\% of feasible plans fail solely due to communicative grounding errors. We demonstrate that a ``Pull-based'' protocol (active querying) is significantly more robust than standard ``Push-based'' instruction, with successful episodes featuring 2x the frequency of clarification requests. This research isolates the mechanism of \textit{active uncertainty reduction} as a prerequisite for safe human-AI and robot-robot collaboration.
\end{abstract}
\section{Introduction}
\label{sec:introduction}

While Large Language Models (LLMs) demonstrate remarkable symbolic reasoning, they effectively operate as ``brains in a jar.'' Trained on vast static datasets—a paradigm often termed ``Internet AI'' \cite{duan2022surveyembodiedaisimulators}—these models lack the proprioceptive feedback loops required to understand physical constraints. They can describe a kitchen, but they cannot inherently reason about who can see the fridge and who cannot \cite{ahn2022saycan, zhao2025foundation}.

This limitation is not merely physical; it is social. In collaborative environments, this disembodiment creates a critical failure mode: the inability to model a partner's perspective. When a knowledgeable agent assumes its partner shares its global view, it succumbs to \textit{Privileged Information Bias}, or the ``Curse of Knowledge'' \cite{camerer1989curse}. Despite the push toward Embodied AI \cite{ramrakhya2025groundingmultimodalllmsembodied, mecattaf2025littleconversationlittleaction}, current frameworks often assume homogeneous agents with shared perception, neglecting the friction that arises when agents must negotiate distinct realities \cite{du2025constrainedhumanaicooperationinclusive}.

We address this gap by embedding a pre-trained LLM into a Leader-Follower dyad within the AI2-THOR simulation environment \cite{kolve2017ai2thor}. We introduce an Asymmetric Assistive Reasoning task where a ``Leader'' with full vision must guide a ``Follower'' with severe visual impairments. This asymmetry forces the LLM to negotiate information gaps rather than simply plan actions. To measure the impact of communication protocols, we contrast a standard ``Push-based'' instruction model against a ``Pull-based'' active querying protocol.

Our experiments reveal a stark ``Success Gap.'' While the Leader agent successfully perceives and navigates to targets in 35.0\% of episodes, the collaborative team succeeds only 17.0\% of the time. This 18-point drop indicates that nearly half of all feasible plans fail solely due to communicative grounding errors. We demonstrate that these failures stem from open-loop instruction, where the Leader issues commands based on an egocentric view (e.g., ``turn left'' relative to itself). By shifting to a ``Pull-based'' protocol, where the Follower actively flags ambiguities, the team mitigates the Curse of Knowledge and restores performance.

This work makes the following contributions:
\begin{itemize}
    \item \textbf{Quantifying the Gap:} We identify a quantitative ``Success Gap,'' establishing that 50\% of feasible navigation plans fail due to grounding errors rather than execution capability.
    \item \textbf{Protocol Evaluation:} We demonstrate that a ``Pull-based'' active querying protocol significantly outperforms open-loop instruction in resolving Privileged Information Bias.
    \item \textbf{Framework:} We provide a reproducible Asymmetric Assistive Reasoning framework for evaluating Theory of Mind and perspective-taking in embodied LLMs.
\end{itemize}

\textbf{Broader Impacts:} Beyond simulation, this research highlights a necessary evolution for assistive robotics. For autonomous systems to safely collaborate with human users—who possess distinct perceptual and physical constraints—they must move beyond blind obedience and develop the capacity to recognize and resolve information asymmetry.
\section{Related Work}
\label{sec:related_work}

Research related to this work addresses the convergence of Embodied AI, Multi-Agent Collaboration, and the specific cognitive failures of Large Language Models (LLMs) when grounded in asymmetric physical realities.

\subsection{The Grounding Problem in AI}
The inability of LLMs to robustly reason about physical dynamics is classically framed as the symbol grounding problem \cite{harnad1990symbol}. While LLMs excel in processing symbolic sequences, their reliance on static datasets leaves them fundamentally disembodied \cite{ahn2022saycan, zhao2025foundation}. This limitation prevents them from intrinsically connecting linguistic tokens with physical actions or understanding cause-and-effect relationships.

Recent critiques suggest that this disembodiment leads to a "superficial" Theory of Mind, where models match behavioral patterns without maintaining a coherent internal world model \cite{trott2025reevaluating}. To bridge this, recent efforts have attempted to augment LLMs with grounding mechanisms that connect abstract language to sensorimotor experiences \cite{lake2017building}, yet these approaches often treat grounding as a single-agent perception task rather than a social negotiation.

\subsection{Embodied AI and 3D Simulation Platforms}
To address the shortcomings of "Internet AI" \cite{duan2022surveyembodiedaisimulators}, the embodied paradigm emphasizes evaluation within interactive environments. Platforms such as AI2-THOR \cite{kolve2017ai2thor}, Habitat \cite{savva2019habitat}, and VirtualHome \cite{puig2018virtualhome} enable agents to practice navigation and manipulation.

However, these platforms traditionally focus on single-agent physics. While recent World Foundation Models (WFMs) like Google's Genie \cite{bruce2024genie} and Meta's V-JEPA 2 \cite{assran2025vjepa2} allow agents to predict temporal dynamics, they predominantly model physical causality (e.g., "if I drop this, it falls") rather than social causality (e.g., "if I say this, will my partner understand?"). Our work specifically targets this communication layer, evaluating how agents resolve ambiguity when their physical world models diverge.

\subsection{Multi-Agent Collaboration and Asymmetry}
LLMs have increasingly served as the cognitive core for Multi-Agent Systems (MAS) \cite{ferrag2025reviewagents, wang2023surveyagents}. Recent frameworks like CoELA \cite{zhang2024building} and ProAgent \cite{zhang2025cut} have demonstrated that LLMs can coordinate decentralized control in complex environments. Similarly, Sun et al. \cite{sun2025collabovercooked} benchmarked LLMs in \textit{Overcooked-AI}, identifying significant deficits in active collaboration.

A critical limitation in frameworks such as CaPo \cite{liu2025capo} and AdaTAMP \cite{liu2025adatamp} is the assumption of homogeneous agents with shared global views. In contrast, our work addresses \textit{asymmetric} collaboration, where agents must resolve "Belief State Divergence" arising from unequal sensor horizons. This creates a risk of "Bias Reinforcement," where unchecked dialogue amplifies errors rather than correcting them \cite{zhang2025bias}. While Patania et al. \cite{patania2025perspact} recently emphasized "Pull-based" interaction to resolve such ambiguities, we extend this by operationalizing it within a strictly sensor-limited dyad to quantify the exact "Success Gap" caused by the Curse of Knowledge.

\subsection{Theory of Mind in Robotics}
Effective communication requires simulating the perspective of less-informed partners—a capability often absent in LLMs \cite{li2025theoryofmind}. In robotics, this manifests as the "Curse of Knowledge," where a planner assumes its instructions are universally grounded \cite{camerer1989curse}. 

Recent studies attempt to minimize this bias via "Devil's Advocate" agents \cite{lee2025conversational} or diverse agent personalities \cite{hsu2025groupthink}. However, \citet{li2025theoryofmind} note that LLMs frequently "hallucinate" shared knowledge in long-horizon tasks. Our framework applies these theories to embodied spatial reasoning. Rather than relying on abstract personalities, we utilize an Advocate-Critic loop to mechanically force uncertainty reduction through active querying, converting the abstract theory of mind failure into a measurable navigation cost.
\section{Approach}
\label{sec:approach}

We formalize the problem of asymmetric collaboration as a partially observable multi-agent pathfinding task where communication is the only bridge between a global planner and a local executor. Unlike standard multi-agent reinforcement learning (MARL) setups that rely on shared reward gradients, our framework utilizes a centralized Large Language Model (LLM) to simulate two distinct cognitive processes—a ``Leader'' and a ``Follower''—within a shared context window.

\begin{figure}[t!]
    \centering
    \includegraphics[width=\columnwidth]{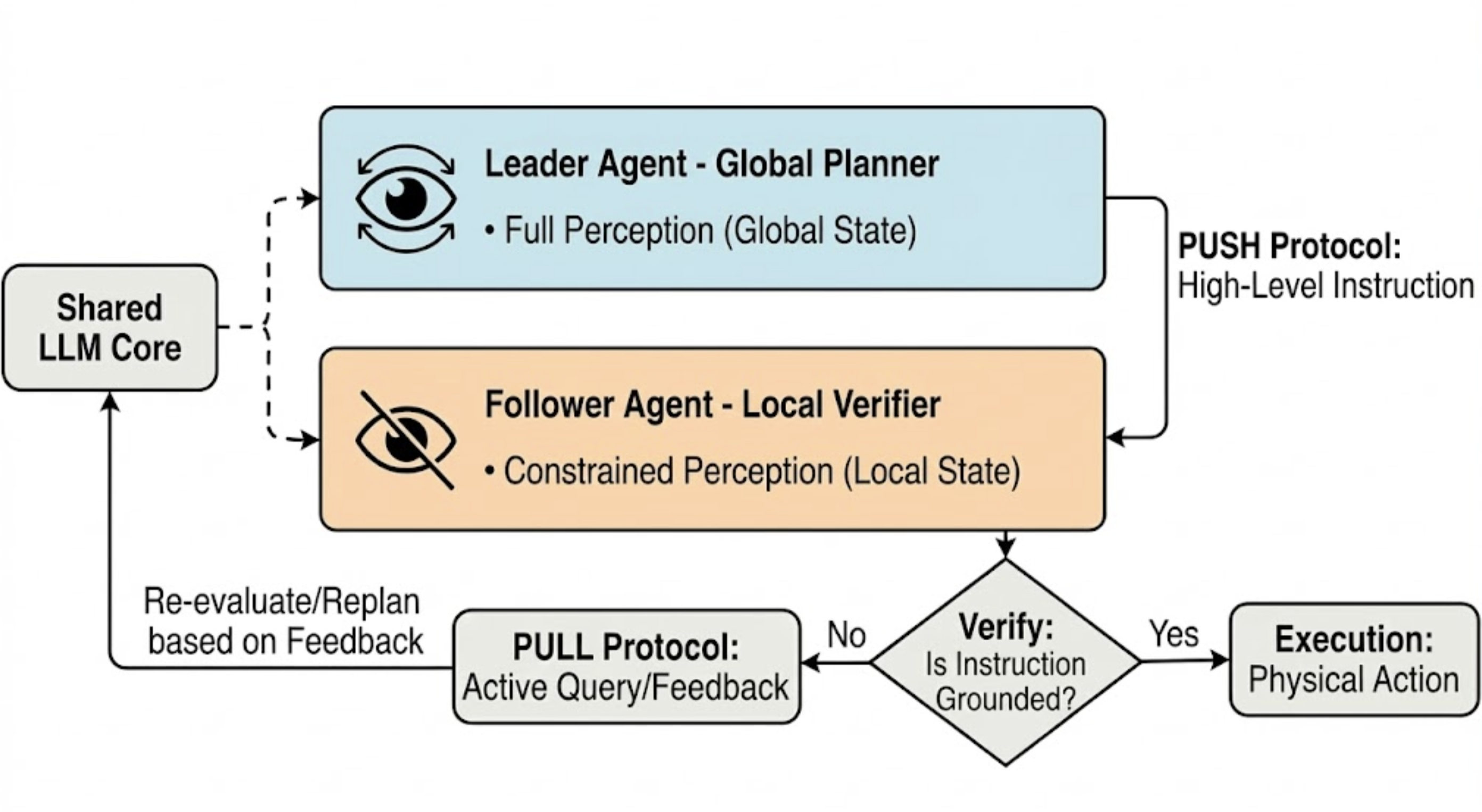}
    \caption{\textbf{The Leader-Follower Architecture.} The Leader utilizes global perception ($S_L$) to ``push'' instructions, while the Follower utilizes local verification ($S_F$) to ``pull'' clarification. The shared LLM core iteratively generates the internal monologue for both roles.}
    \label{fig:workflow}
\end{figure}

\subsection{Problem Formulation}
We define a dyad consisting of two agents, $A_{L}$ (Leader) and $A_{F}$ (Follower), operating in a shared environment $\mathcal{E}$. The agents share a high-level goal $G$ (e.g., ``Find the Apple''), but possess distinct observability constraints:

\begin{itemize}
    \item \textbf{Leader State ($S_{L}$):} The Leader has access to the ground-truth state of the environment. This includes a semantic map of all objects $O = \{o_1, o_2, ..., o_n\}$ and their exact global coordinates $(x, y, z)$. The Leader acts as the \textit{Global Planner}, generating high-level waypoints based on $S_{L}$.
    \item \textbf{Follower State ($S_{F}$):} The Follower operates under severe perceptual constraints. Its observation is limited to an egocentric subset $O_{local} \subset O$, containing only objects within a radius $d_{view} \leq 2.0$m and a field-of-view $\theta = 90^{\circ}$. The Follower acts as the \textit{Local Verifier}, executing motor commands ($M_{step}$, $M_{rotate}$) and validating plan feasibility against $S_{F}$.
\end{itemize}

\subsection{Architecture: Single-Core, Dual-Persona}
To simulate the interaction without the latency of multi-model orchestration, we employ a Single-Core, Dual-Persona architecture. We utilize a single Gemini 2.5 Flash kernel that iteratively generates the internal monologue and external dialogue for both agents. The system prompt enforces a strict separation of knowledge: the model is explicitly forbidden from allowing the Follower persona to access the Leader's global state $S_{L}$.

At each timestep $t$, the system constructs the context as:
\begin{equation}
    C_t = [P_{sys}, H_{dial}, S_{L}^{(t)}, S_{F}^{(t)}]
\end{equation}
Where $P_{sys}$ is the role-defining system prompt and $H_{dial}$ is the shared conversation history. The model then outputs a tuple $(I_{L}, A_{F})$, where $I_{L}$ is the Leader's natural language instruction and $A_{F}$ is the Follower's response (either a physical action or a text query).

\subsection{Interaction Protocols: Push vs. Pull}
We explicitly contrast two interaction modalities to measure the impact of active uncertainty reduction.

\subsubsection{The Push Protocol (Open-Loop)}
In the ``Push'' condition (Method A), the interaction is unidirectional. The Leader broadcasts an instruction $I_{push}$ based on its global state $S_{L}$. The Follower attempts to map $I_{push}$ directly to a physical action $a \in \mathcal{A}$. 
\begin{equation}
    S_{L} \rightarrow I_{push} \rightarrow A_{F}(execute)
\end{equation}
This mimics standard instruction-following baselines where the agent suffers from \textit{Egocentric Bias} \cite{camerer1989curse}, assuming the instructor's perspective is universally valid.

\subsubsection{The Pull Protocol (Closed-Loop)}
In the ``Pull'' condition (Method B), the Follower utilizes a Verification Module to check $I_{push}$ against its local constraints $S_{F}$. If the instruction references an ungrounded landmark (e.g., ``Go to the sofa'' when no sofa is visible in $S_{F}$), the Follower triggers a ``Pull'' query $Q_{pull}$.

\begin{equation}
\begin{split}
    S_{L} \rightarrow I_{push} & \rightarrow \text{Verify}(S_{F}) \\
    & \xrightarrow{fail} Q_{pull} \rightarrow S_{L}(\text{Re-Ground})
\end{split}
\end{equation}

This closes the loop, forcing the Leader to translate global coordinates into local, relative cues (e.g., ``Turn right 90 degrees'').

\subsection{Experimental Testbed}
The framework is implemented in AI2-THOR \cite{kolve2017ai2thor} using the ManipulaTHOR \cite{ehsani2021manipulathorframeworkvisualobject} asset subset. We enforce the asymmetry by ray-casting from the Follower's camera and filtering the returned object list to exclude any entity beyond the $2.0$m horizon.

\subsection{Algorithm}
Algorithm \ref{alg:framework} formalizes this interaction. In the default ``Push'' state, the Leader broadcasts plans based on global perception ($e_{global}$). The critical deviation occurs when the Follower's local verification fails ($v_{check}$ is ungrounded). This triggers the ``Pull'' branch, where a query $Q$ is fed back into the Leader's planner ($\mathcal{M}_{\mathrm{adv}}$).

\begin{algorithm}[h!]
\caption{Asymmetric Leader-Follower Framework}
\label{alg:framework}
\SetAlgoLined 
\KwIn{Global state $e_{global}$, Local state $e_{local}$, goal $g$}
\KwOut{Executed actions $A$}

\While{Goal $g$ is not achieved}{
  \textbf{Leader:} Perceive global state $e_{global}$ and ``Push'' initial plan
  $I_{push} = \mathcal{M}_{\mathrm{adv}}(e_{global}, g)$\;
  
  \textbf{Follower:} Verify $I_{push}$ against local constraints $e_{local}$
  $v_{check} = \mathcal{M}_{\mathrm{crit}}(I_{push}, e_{local})$\;
  
  \eIf{$v_{check}$ is grounded}{
    \textbf{Execution:} Translate instruction for execution
    $A = \mathcal{M}_{\mathrm{exec}}(I_{push})$\;
  }{
    \textbf{Pull Protocol:} Generate query $Q$ to communicate local constraints
    $Q = \mathcal{M}_{\mathrm{query}}(I_{push}, e_{local})$\;
    \textbf{Re-Grounding:} Leader revises plan based on $Q$
    $I_{push} = \mathcal{M}_{\mathrm{adv}}(e_{global}, g, Q)$\;
  }
}
\Return{$A$}
\end{algorithm}
\section{Experiments}
\label{sec:experiments}

We designed an experiment to isolate the specific benefits of shared perception and guided instruction, comparing our Asymmetric Leader-Follower model against single-agent baselines in a controlled, replicable environment.

\begin{figure*}[t!]
    \centering
    \includegraphics[width=\textwidth]{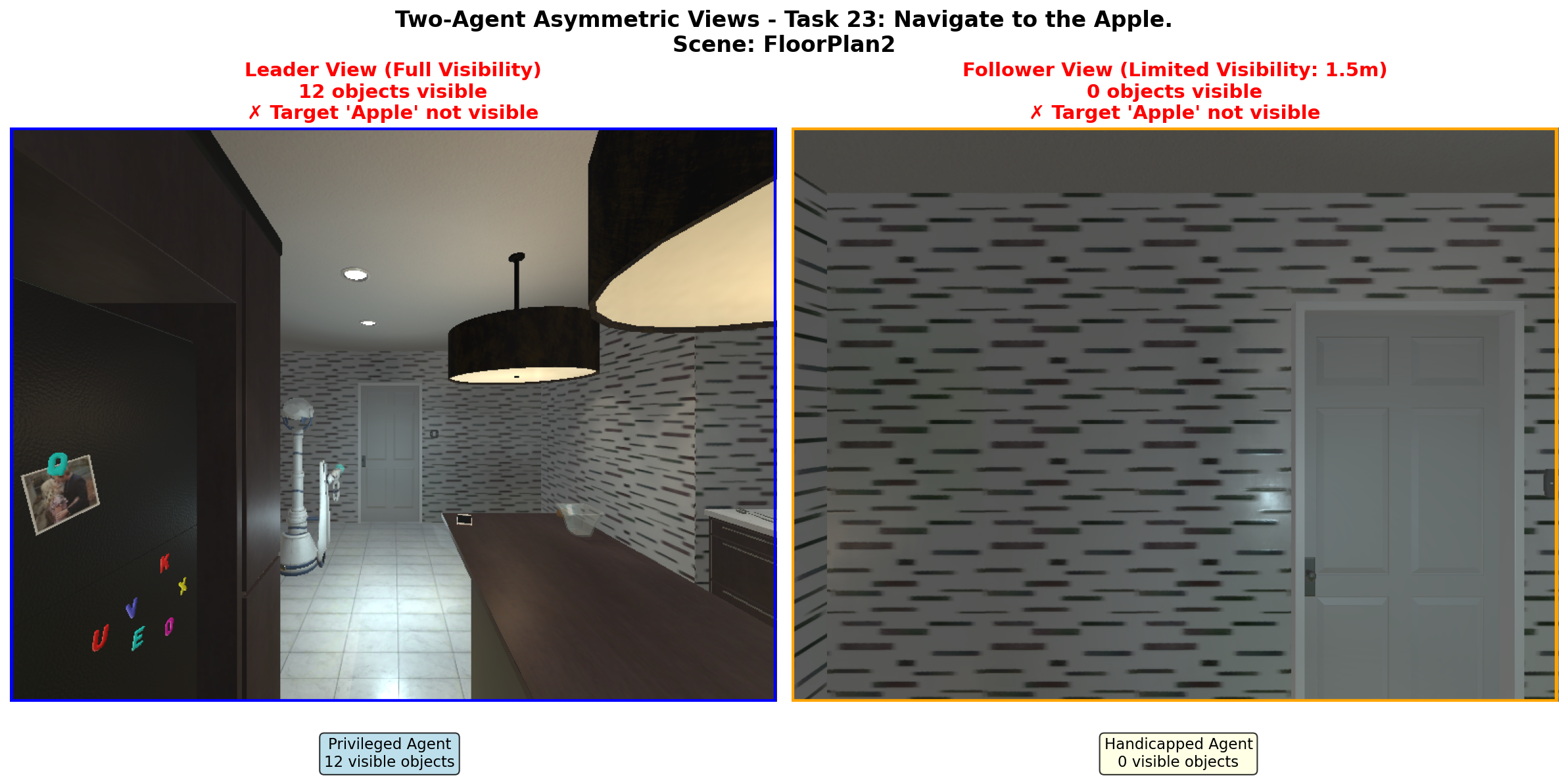}
    \caption{\textbf{Visualizing the Information Asymmetry (Task 23).} The \textbf{Leader} (left) perceives the full scene geometry (12 objects visible), identifying the target ``Apple'' relative to the room layout. The \textbf{Follower} (right) operates under a 2.0m visibility handicap (0 objects visible), seeing only a blank wall. This discrepancy creates the ``Curse of Knowledge,'' where the Leader must infer the Follower's blindness to provide effective guidance.}
    \label{fig:asymmetry}
\end{figure*}

\subsection{Testbed and Task Design}
We utilize the AI2-THOR simulation platform \cite{kolve2017ai2thor}, specifically the ManipulaTHOR subset \cite{ehsani2021manipulathorframeworkvisualobject}, chosen for its high-fidelity physics and complex indoor clutter.

The task is Object-Goal Navigation: The agent must navigate from a random spawn point $P_{start}$ to a target object $O_{target}$ (e.g., ``Find the Apple'').

\subsubsection{Data Generation Pipeline}
To ensure task validity (addressing the ``impossible target'' problem), we implemented a rigorous preprocessing pipeline using the environment's `GetReachablePositions` API. We generated 1,320 candidate episodes and filtered them to enforce two constraints:
\begin{enumerate}
    \item \textbf{Reachability:} A valid nav-mesh path must exist between $P_{start}$ and $O_{target}$.
    \item \textbf{Non-Triviality:} The geodesic distance must be $d > 1.5$m to prevent the agent from spawning immediately in front of the goal.
\end{enumerate}
From this filtered set, we randomly sampled a fixed \textbf{Benchmark Set} of 100 tasks across 4 room types (Kitchen, Bathroom, Living Room, Bedroom).

\subsection{Model Conditions}
We evaluate three distinct policy conditions to quantify the ``Collaboration Boost'':
\begin{enumerate}
    \item \textbf{Baseline Agent (Egocentric Control):} A single agent with full sensory perception ($O_{full}$). This represents the \textit{Performance Ceiling}.
    \item \textbf{Handicapped Agent (Sensory Control):} A single agent operating under the Follower's constraints ($O_{partial}$, max view distance 2.0m). This establishes the \textit{Zero-Shot Performance Floor}.
    \item \textbf{Two-Agent Dyad (Ours):} The Leader-Follower framework described in Section \ref{sec:approach}.
\end{enumerate}

\subsubsection{Model Development \& Optimization}
During preliminary development, we observed that larger models (e.g., Gemini Pro) frequently hallucinated object positions in the ``Follower'' role. We resolved this by switching to the `gemini-2.5-flash` variant and enforcing a temperature of $0.0$, which stabilized the coordinate-to-action mapping.

\subsection{Evaluation Metrics}
We report results using standard Embodied AI metrics:
\begin{itemize}
    \item \textbf{Success Rate (SR):} The percentage of episodes where the agent reaches $d \leq 1.0$m from the target. We selected the $1.0$m threshold to align with the standard ALFRED benchmark success criteria \cite{shridhar2020alfredbenchmarkinterpretinggrounded}.
    \item \textbf{Average Steps to Success (STS):} The average number of simulation steps (movement + rotation) taken to reach the goal. \textit{Note:} This metric is calculated \textbf{only over successful episodes} to prevent maximum-length failures ($T_{max}$) from artificially inflating the average, which would obscure efficiency gains.
    \item \textbf{Success weighted by Path Length (SPL):} A strict measure of path efficiency: $\frac{1}{N} \sum_{i=1}^{N} S_i \frac{L_i}{\max(P_i, L_i)}$, where $L_i$ is the optimal path and $P_i$ is the observed path.
    \item \textbf{Collaboration Metrics:} The volume of ``Push'' instructions (Leader) versus ``Pull'' queries (Follower).
\end{itemize}

\subsection{Quantitative Performance}
We executed the full 100-task benchmark across all three conditions. The aggregate performance is summarized in Table \ref{tab:main_results}, and the communication dynamics are detailed in Table \ref{tab:collab_metrics}.

\begin{table}[htbp]
  \centering
  \caption{Aggregate Performance. SPL represents path efficiency (higher is better). STS represents temporal cost (lower is better).}
  \label{tab:main_results}
  \resizebox{\columnwidth}{!}{%
  \begin{tabular}{l c c c}
    \toprule
    \textbf{Policy Condition} & \textbf{Success Rate (SR)} & \textbf{Avg. STS} & \textbf{SPL} \\
    \midrule
    Baseline Agent (Solo) & 16.0\% & 4.44 & 0.14 \\
    Handicapped Agent (Solo) & 11.0\% & 4.36 & 0.09 \\
    \midrule
    Two-Agent (Leader View) & 35.0\% & 12.23 & - \\
    Two-Agent (Follower View) & \textbf{17.0\%} & 7.24 & 0.15 \\
    \bottomrule
  \end{tabular}%
  }
\end{table}

\begin{table}[htbp]
  \centering
  \caption{Communication Analysis: The ``Pull'' protocol (active querying) is the primary driver of success, appearing 2x more frequently in successful episodes.}
  \label{tab:collab_metrics}
  \resizebox{\columnwidth}{!}{%
  \begin{tabular}{l c}
    \toprule
    \textbf{Metric} & \textbf{Count / Episode} \\
    \midrule
    Avg. Leader Instructions (on Success) & 24.41 \\
    Avg. Leader Instructions (on Failure) & 25.99 \\
    \midrule
    Avg. Follower Queries (on Success) & \textbf{2.00} \\
    Avg. Follower Queries (on Failure) & 0.99 \\
    \bottomrule
  \end{tabular}%
  }
\end{table}

\subsection{Diagnostic Analysis}
\label{sec:diagnostic_analysis}

\subsubsection{The Success Gap}
Our results reveal a critical ``Success Gap.'' While the Leader agent successfully perceives the target in 35.0\% of episodes, the collaborative team succeeds only 17.0\% of the time. This \textbf{18-point drop} implies that nearly 50\% of feasible plans fail during transmission. The Leader ``knows'' the path but fails to translate it into a grounded instruction the Follower can verify.

\subsubsection{Mechanism of Success: Push vs. Pull}
Table \ref{tab:collab_metrics} isolates the mechanism of success. In successful episodes, the Follower issued \textbf{2.00 active queries} per episode, compared to just 0.99 in failed episodes. Crucially, the volume of Leader instructions (``Push'') remained constant ($\approx 25$) regardless of outcome. This confirms that success is not driven by \textit{more} instructions, but by \textit{more verification}.

\subsubsection{Qualitative Error Analysis}
To investigate the root cause of the Success Gap, we audited failure trajectories. A representative failure occurred in \textbf{Task 38 (Find the Apple)}:
\begin{itemize}
    \item \textbf{Leader State:} Perceives the Apple on a table 5m away. Issues command: ``Move Forward.''
    \item \textbf{Follower State:} Facing a blank wall (distance 1.5m).
    \item \textbf{Failure Mode:} The Follower obeyed the ``Move Forward'' command blindly (Push protocol), colliding with the wall. In successful runs of similar tasks, the Follower utilized the Pull protocol to ask: ``I see a wall. Which way is the table?'' prompting the Leader to correct with ``Turn Right 90 degrees.''
\end{itemize}
This confirms that failures are rarely due to disobedience, but rather due to \textit{ungrounded obedience} to egocentric instructions.
\section{Analysis}
\label{sec:analysis}

The experimental evaluation of the ``Emergence'' framework isolates the specific friction points where the symbolic reasoning of Large Language Models (LLMs) conflicts with the hard constraints of a physics-rich environment. By embedding the Gemini-2.5-flash model in the AI2Thor environment, ManipulaTHOR, we move beyond measuring abstract reasoning to measuring grounded efficacy. The following analysis deconstructs the performance data stratified across the Baseline, Handicapped, and Two-Agent conditions, diagnosing the specific cognitive and communicative pathologies that persist in collaborative spatial reasoning.

\subsection{Performance Landscape}
Our results establish a clear hierarchy of competence that quantifies both the ``Sensory Tax'' of the handicap and the ``Collaboration Boost'' of the Leader-Follower architecture.

\begin{figure}[h!]
    \centering
    \includegraphics[width=0.5\textwidth]{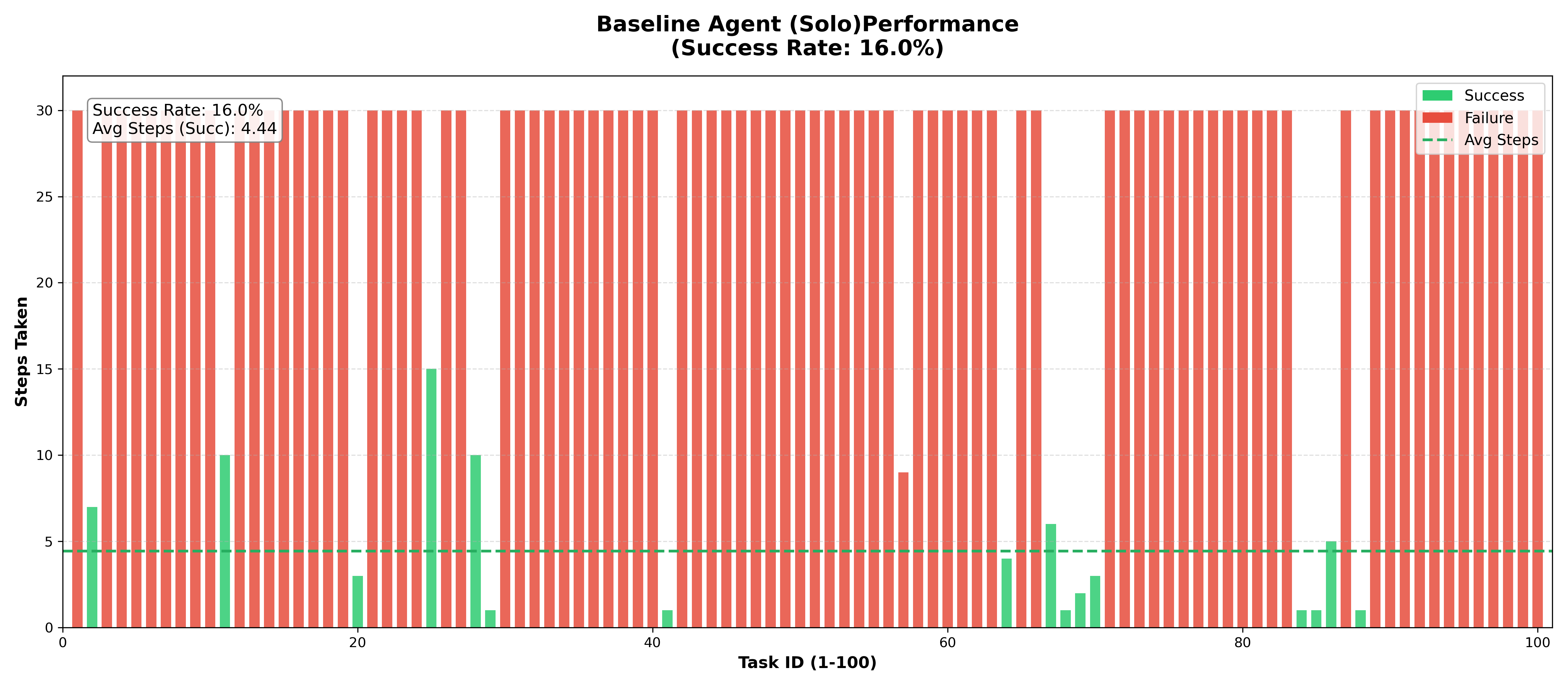}
    \caption{The baseline agent's performance (16.0\% SR) illustrates the ``Zero-Shot Ceiling,'' where success is largely determined by favorable spawn locations rather than systematic search.}
    \label{fig:baseline_perf}
\end{figure}

\begin{figure}[h!]
    \centering
    \includegraphics[width=0.5\textwidth]{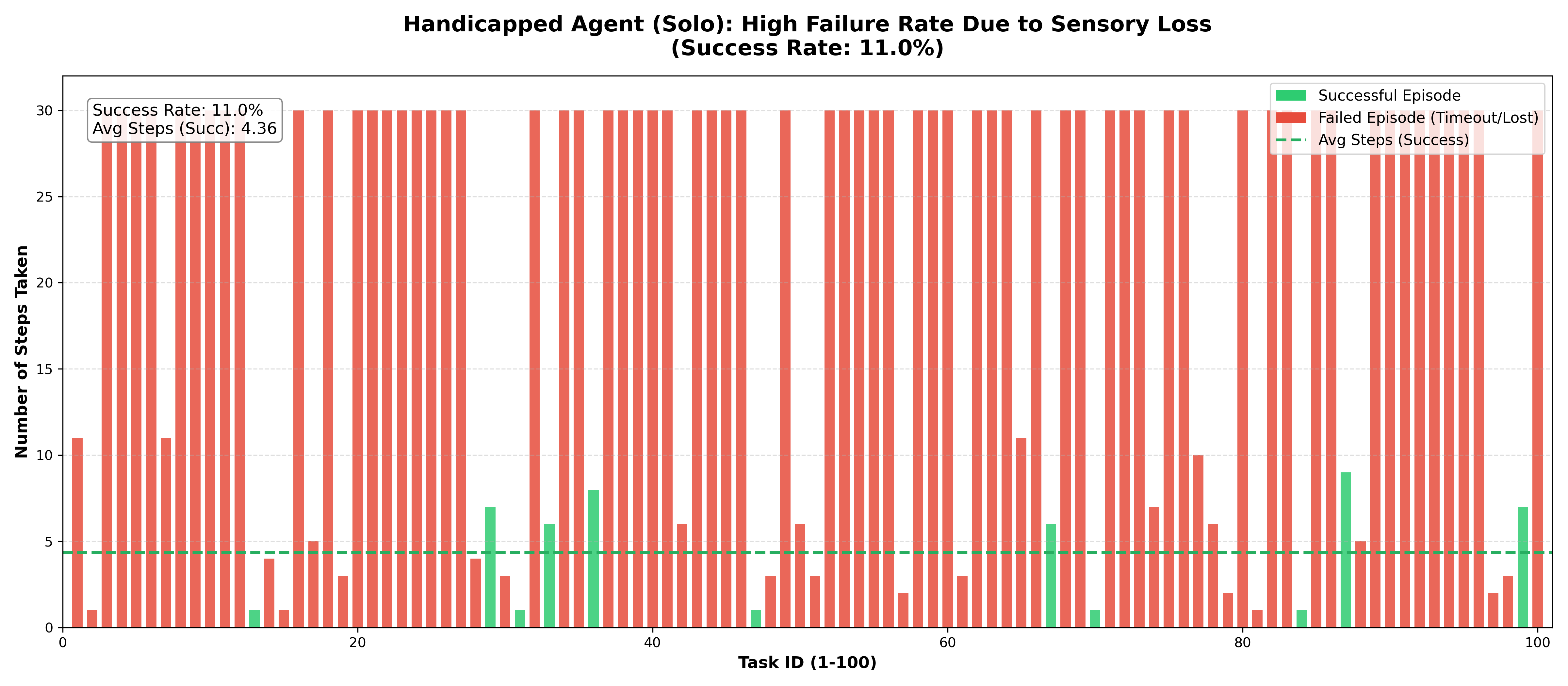}
    \caption{The handicapped agent's performance drop (to 11.0\% SR) quantifies the ``Sensory Tax,'' confirming that semantic reasoning cannot compensate for a lack of distal visual cues.}
    \label{fig:handicapped_perf}
\end{figure}

\subsubsection{The Zero-Shot Ceiling and Semantic-Spatial Dissonance}
The Baseline Agent, equipped with full visual acuity, achieved a Task Success Rate (SR) of 16.0\%. To contextualize this figure, it must be weighed against the broader landscape of Embodied AI. State-of-the-art agents trained through Reinforcement Learning (RL) or Imitation Learning (IL) on AI2-THOR typically achieve a success rate ranging from 26\% to over 70\%, depending on training volume and pre-mapping capabilities \cite{ma2024doze}.
The disparity between our Baseline (16.0\%) and these specialized agents highlights the ``Zero-Shot Penalty.'' Unlike RL agents that build implicit collision policies through millions of trial-and-error steps, the LLM agent relies entirely on semantic reasoning. The data suggest a \textbf{Semantic-Spatial Dissonance}: the agent possesses the semantic knowledge to identify a ``refrigerator'' but lacks the procedural ``proprioception'' to navigate the coordinate-level sequence required to reach it. The agent effectively operates as a ``brain in a jar,'' translating high-level intent into low-level motor commands without a learned intuition for the environment's geometry.

\subsubsection{Quantifying the Sensory Tax}
The Handicapped Agent condition simulated ``reduced perception'' by restricting visibility to 2.0 meters. This handicap induced a measurable performance degradation, dropping the SR from 16.0\% to 11.0\%. This 31.3\% relative decline represents the ``Sensory Tax''—the quantifiable cost of losing distal visual cues. Without the ability to spot landmarks, like couches or counters across the room, the agent was reduced to a stochastic local search, confirming that semantic intent cannot fully compensate for perceptual blindness.

\begin{figure}[h!]
    \centering
    \includegraphics[width=0.5\textwidth]{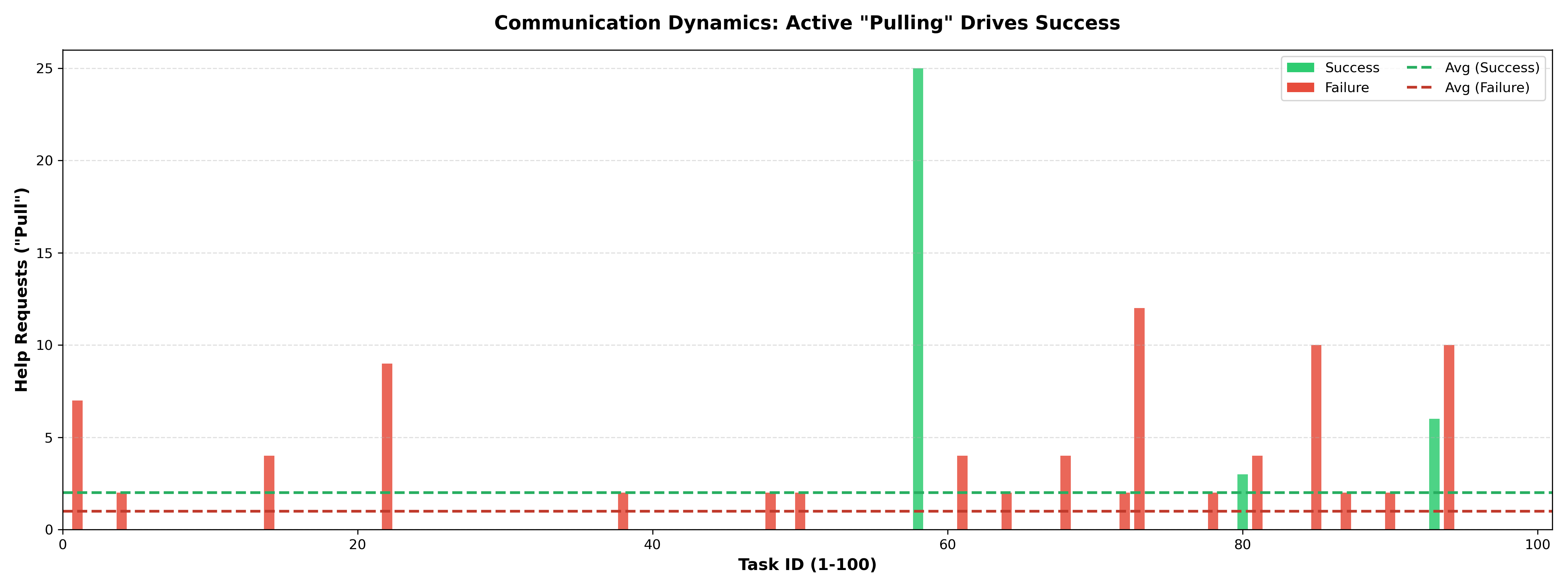}
    \caption{The impact of active querying: Successful episodes (Green) feature 2x the frequency of ``Pull'' requests compared to failed episodes (Red), validating the ``Push-Pull'' hypothesis.}
    \label{fig:collab_boost}
\end{figure}

\subsubsection{The Collaboration Boost}
The primary hypothesis was validated: the Two-Agent system successfully mitigated the sensory handicap. The Assisted Handicapped agent achieved an SR of 17.0\%, recovering and slightly exceeding the performance of the fully sighted Baseline (16.0\%). This 54.5\% improvement over the solo handicapped condition confirms that the Leader successfully transferred spatial knowledge to the Follower, effectively acting as a remote sensory organ. However, the fact that the pair only matched the solo baseline suggests that collaboration is \textit{restorative} rather than \textit{additive}: it heals the disability but does not yet yield super-human performance.

\subsection{Error Analysis}
While the aggregate metrics show success, a deeper dive into the dyadic performance reveals a critical failure mode, which we term the ``Success Gap.''

\subsubsection{The Leader-Follower Disconnect}
A stark discrepancy exists between the Leader's perception and the Follower's execution:
\begin{itemize}
    \item \textbf{Leader Success Rate:} 35.0\%
    \item \textbf{Follower (Assisted) Success Rate:} 17.0\%
    \item \textbf{The Gap:} 18.0 percentage points
\end{itemize}
In 35.0\% of the episodes, the Leader agent successfully identified the target and navigated to it. Yet, in more than half of those successful instances, it failed to guide the Handicapped partner to the same destination. The 18-point gap signifies a failure in Theory of Mind (ToM). The Leader suffers from the ``Curse of Knowledge''; it perceives the target (e.g., ``Red Mug, 5m'') and fails to simulate the belief state of the Follower, who perceives only ``Wall, 1.5m.'' Consequently, the Leader issues instructions that are accurately grounded in its own reality but referentially ambiguous in the partner's reality \cite{patania2025perspact}.

\subsubsection{The Push-Pull Hypothesis and the Price of Passivity}
The communication logs provide the causal mechanism for this gap. We observed a constant volume of ``Push'' communication from the Leader ($\approx$25 instructions/episode) across both success and failure cases. ``Pushing'' more instructions did not correlate with success.

Instead, success was entirely dependent on the ``Pull'' mechanism:
\begin{itemize}
    \item \textbf{Help Requests (Successful Episodes):} 2.00 per episode
    \item \textbf{Help Requests (Failed Episodes):} 0.99 per episode
\end{itemize}
The 18.0\% gap represents the ``Price of Passivity.'' The 2:1 ratio in help requests indicates that the ``Local Verifier'' (Follower) is the linchpin of the architecture. In failed episodes, the low querying rate suggests the Follower failed to recognize its own divergence from the plan, treating the Leader's instructions as absolute truths rather than hypotheses requiring local verification. When the ``Pull'' mechanism is dormant, the dyad reverts to a naive ``Blind Leading the Blind'' topology, where the Follower executes instructions that are physically impossible in its local frame.

\subsubsection{Qualitative Analysis: The Anatomy of Miscommunication}
To determine if the 18.0\% Success Gap resulted from the Follower ignoring commands or the Leader issuing ungrounded commands, we audited the conversation logs of failed episodes. We found that ``Compliance Failure'' (Follower ignoring a valid command) was rare. The predominant failure mode was ``Blind Obedience to Hallucinated Relativity.''

As illustrated in Table \ref{tab:log_analysis}, the Leader frequently provides relative directions (e.g., ``to your left'') based on its \textit{own} camera orientation, which differs from the Follower's. Without an active ``Pull'' query from the Follower to verify the reference frame, the Follower executes the command faithfully but incorrectly.

\textbf{Root Cause Analysis (Task 38):} In a representative failure case, the Leader perceived the Apple at coordinates $(1.5, 0.5, 3.0)$ and issued the command ``Move Left.'' The Follower, located at $(1.5, 0.5, 1.5)$ and facing a wall, interpreted ``Left'' relative to its own orientation. This resulted in a collision. In contrast, successful dyads utilized the Pull mechanism to ask: ``Which way is the Apple relative to the wall?'' This query forced the Leader to perform \textbf{Frame Switching}, translating the egocentric instruction into an allocentric or landmark-based cue (e.g., ``Turn 90 degrees right away from the wall''). This confirms that the ``Pull'' mechanism enables success not just by re-grounding, but by forcing the Leader to abandon its egocentric reference frame.

\begin{table}[h!]
\small
\centering
\resizebox{\columnwidth}{!}{%
\begin{tabular}{l|p{6cm}} 
\hline
\textbf{Metric} & \textbf{Log Entry Example} \\ \hline
Step & 12 \\
Leader View & Target (Apple) visible at [x: 1.5, y: 0.5, z: 3.0] \\
Follower View & Obstructed (Wall) \\
Leader Instruction & ``The apple is directly to your left. Move Left.'' \\
\textbf{Analysis} & \textit{Incorrect Grounding. The 'Left' is relative to the Leader's start position, not the Follower's current orientation.} \\
Follower Response & ``Moving Left.'' \\
Result & Collision with Wall (Episode Fail) \\ \hline
\end{tabular}%
}
\caption{An example of ``Privileged Information Bias'' leading to Navigation Failure. The Follower obeys the command, but the command is ungrounded in the Follower's frame of reference.}
\label{tab:log_analysis}
\end{table}

\subsection{Ablation Study: Temporal Horizon and Stability}
\label{sec:ablation}

To rigorously test the hypothesis that the 30-step limit ($T_{max}=30$) artificially constrained performance, we conducted an ablation study by re-running 91 failed tasks with the horizon extended to $T_{max}=60$.

\begin{table}[h!]
\centering
\resizebox{\columnwidth}{!}{%
\begin{tabular}{l|c|c|c}
\hline
\textbf{Agent} & \textbf{30-Step SR} & \textbf{60-Step SR} & \textbf{Relative Imp.} \\ \hline
Leader & 28.6\% & 34.1\% & +19.2\% \\
Handicapped & 8.8\% & 14.3\% & +62.5\% \\ \hline
\end{tabular}%
}
\caption{Results of the 60-step Ablation Study showing performance recovery.}
\label{tab:ablation_results}
\end{table}

As shown in Table \ref{tab:ablation_results}, relaxing the temporal constraint resulted in a significant relative improvement, particularly for the Handicapped agent (+62.5\%). This confirms that a subset of ``failed'' plans were viable but required longer horizons to converge.

However, a granular analysis of the recovered tasks reveals that simple ``horizon truncation'' is not the sole factor. Among the 7 newly successful Handicapped episodes, only 1 (14.3\%) actually required more than 30 steps. The remaining 6 succeeded in $<30$ steps during the re-run. This indicates that the performance boost is largely attributable to stochastic robustness. The extended horizon allows the agent more opportunities to recover from initial ``lucky stumbles'' or bad random seeds, effectively smoothing out the variance inherent in zero-shot LLM navigation.

\subsection{Discussion}

\subsubsection{Limitations and Feasibility Analysis}
The Leader agent achieved a success rate of 35.0\%. However, attributing the remaining 65\% solely to reasoning failures is a mischaracterization of the results due to two critical factors:

\paragraph{Undefined Feasibility Ceiling:}
Unlike standard benchmarks that utilize human baselines to establish a ``perfect play'' ceiling (100\%), this study relies on zero-shot LLM performance. Given the complexity of the ManipulaTHOR environments, it is statistically probable that a subset of targets are unreachable within 30 steps regardless of intelligence (due to spawn distance). Without a human baseline to establish that 100\% of these tasks are solvable in $<30$ steps, the Leader's 35\% success rate should be viewed as a lower bound of capability, not an absolute ceiling.

\paragraph{Semantic Random Walks:}
Current ``post-hoc embodiment'' relies on the LLM's context window to store history. The text-based serialization results in perception loss, flattening 3D geometry into a list of items. The agent struggles to build a coherent ``mental map,'' leading to ``semantic random walks'' where it revisits invalid locations. As evidenced by our ablation study, while extending the episode length improves success, it does not fundamentally resolve the lack of spatial memory.

\subsubsection{Future Direction}
The analysis of the Push-Pull dynamic suggests multi-agent architectures should move beyond ``cooperation'' to ``incentivized uncertainty reduction.'' The current prompt likely biased agents toward agreeableness. To bridge the gap, the ``Follower'' agent should be explicitly incentivized to reject ambiguous instructions. We propose a ``Devil's Advocate'' reward function for future experiments, where the Follower is rewarded not just for reaching the goal, but for identifying and flagging uncertainties in the Leader's plan \cite{lee2025conversational}. True embodied intelligence emerges not when agents agree, but when they successfully resolve their disagreement.
\section{Conclusion}
\label{sec:conclusion}

\subsection{Summary of Contributions}
This work systematically evaluated the ``Grounding Gap'' in asymmetric collaboration. \textbf{By introducing a reproducible Asymmetric Assistive Reasoning testbed,} we quantified the specific cost of the \textit{Privileged Information Bias} in LLM-driven agents. Our experiments demonstrated that raw semantic intelligence does not guarantee collaborative success: despite the Leader identifying targets in 35.0\% of episodes, the team failed to execute plans nearly half the time (17.0\% success).

Our analysis attributes this failure to a lack of Theory of Mind, specifically the inability of the Leader to simulate the Follower's sensory constraints. Crucially, we identified the causal mechanism for resolution: ``Push-based'' broadcasting fails to resolve ambiguity, whereas ``Pull-based'' active querying restores performance. Successful episodes were characterized by a 2:1 ratio of Follower queries to failures, confirming that embodied intelligence emerges not from blind obedience, but from the active negotiation of belief states.

\subsection{Real World Implications}
These findings have direct implications for Assistive and Task-Oriented Robotics. Real-world systems invariably function under asymmetric perception—whether due to sensor occlusion, latency, or distinct physical vantage points. Our results suggest that current ``instruction-following'' paradigms are insufficient for safety-critical tasks. To mitigate dangerous collisions or execution failures, autonomous systems must be designed with \textit{Epistemic Anxiety}—the ability to recognize when an instruction is ungrounded and the agency to pause and query the human or supervisor before acting.

\subsection{Future Work}
These findings motivate three specific shifts in multi-agent system design:

\begin{description}
    \item[The ``Devil's Advocate'' Objective:] To formalize the ``Pull'' mechanism, future training objectives should explicitly reward agents for questioning ambiguous instructions rather than maximizing agreeableness. This moves active querying from an emergent behavior to a learned policy.
    
    \item[Dynamic State Synchronization:] We intend to expand this framework to dynamic environments with moving obstacles. In such settings, the ``Success Gap'' will likely exacerbate, requiring real-time belief synchronization beyond static navigation.
    
    \item[Visual-Language Integration:] Finally, we aim to reduce the ``Sensory Tax'' by integrating Vision-Language Models (VLMs) capable of sharing image patches. This would allow the Leader to transmit a ``visual imagination'' to the blind follower, bridging the gap between semantic description and sensorimotor reality.
\end{description}
\section*{Acknowledgments}
We extend our gratitude to our course instructor, Professor Jinho Choi, and our teaching assistant, Grace Byun. Their guidance on the experimental design and their feedback on the framing of the ``Success Gap'' were instrumental in shaping this research.

\bibliography{custom}

@misc{duan2022surveyembodiedaisimulators,
    title         = {A Survey of Embodied AI: From Simulators to Research Tasks}, 
    author        = {Jiafei Duan and Samson Yu and Hui Li Tan and Hongyuan Zhu and Cheston Tan},
    year          = {2022},
    eprint        = {2103.04918},
    archivePrefix = {arXiv},
    primaryClass  = {cs.AI},
    url           = {https://arxiv.org/abs/2103.04918} 
}

@article{zhao2025foundation,
  title={Foundation Model Driven Robotics: A Comprehensive Review},
  author={Khan, Muhammad Tayyab and Waheed, Ammar},
  journal={arXiv preprint arXiv:2507.10087},
  year={2025},
  url={https://arxiv.org/abs/2507.10087}
}

@article{harnad1990symbol,
    author  = {Stevan Harnad},
    title   = {The symbol grounding problem},
    journal = {Physica D: Nonlinear Phenomena},
    year    = {1990},
    volume  = {42},
    number  = {1--3},
    pages   = {335--346}
}

@misc{kolve2017ai2thor,
    title         = {{AI2-THOR}: An Interactive {3D} Environment for Visual {AI}},
    author        = {Eric Kolve and Roozbeh Mottaghi and Winson Han and Eli VanderBilt and Luca Weihs and Alvaro Herrasti and Daniel Gordon and Yuke Zhu and Abhinav Gupta and Ali Farhadi},
    year          = {2017},
    eprint        = {1712.05474},
    archivePrefix = {arXiv},
    primaryClass  = {cs.CV},
    url           = {https://arxiv.org/abs/1712.05474}
}

@article{savva2019habitat,
  title={Habitat: A Platform for Embodied AI Research},
  author={Manolis Savva and Abhishek Kadian and Oleksandr Maksymets and Yili Zhao and Erik Wijmans and Bhavana Jain and Julian Straub and Jia Liu and Vladlen Koltun and Jitendra Malik and Devi Parikh and Dhruv Batra},
  journal={2019 IEEE/CVF International Conference on Computer Vision (ICCV)},
  year={2019},
  pages={9338-9346},
  url={https://api.semanticscholar.org/CorpusID:91184540}
}

@article{puig2018virtualhome,
  title={VirtualHome: Simulating Household Activities Via Programs},
  author={Xavier Puig and Kevin Kyunghwan Ra and Marko Boben and Jiaman Li and Tingwu Wang and Sanja Fidler and Antonio Torralba},
  journal={2018 IEEE/CVF Conference on Computer Vision and Pattern Recognition},
  year={2018},
  pages={8494-8502},
  url={https://api.semanticscholar.org/CorpusID:49317780}
}

@article{lake2017building,
    author  = {Brenden M. Lake and Tomer D. Ullman and Joshua B. Tenenbaum and Samuel J. Gershman},
    title   = {Building machines that learn and think like people},
    journal = {Behavioral and Brain Sciences},
    year    = {2017},
    volume  = {40},
    pages   = {e253},
    doi     = {10.1017/S0140525X16001837}
}

@misc{ahn2022saycan,
      title={Do As I Can, Not As I Say: Grounding Language in Robotic Affordances}, 
      author={Michael Ahn and Anthony Brohan and Noah Brown and Yevgen Chebotar and Omar Cortes and Byron David and Chelsea Finn and Chuyuan Fu and Keerthana Gopalakrishnan and Karol Hausman and Alex Herzog and Daniel Ho and Jasmine Hsu and Julian Ibarz and Brian Ichter and Alex Irpan and Eric Jang and Rosario Jauregui Ruano and Kyle Jeffrey and Sally Jesmonth and Nikhil J Joshi and Ryan Julian and Dmitry Kalashnikov and Yuheng Kuang and Kuang-Huei Lee and Sergey Levine and Yao Lu and Linda Luu and Carolina Parada and Peter Pastor and Jornell Quiambao and Kanishka Rao and Jarek Rettinghouse and Diego Reyes and Pierre Sermanet and Nicolas Sievers and Clayton Tan and Alexander Toshev and Vincent Vanhoucke and Fei Xia and Ted Xiao and Peng Xu and Sichun Xu and Mengyuan Yan and Andy Zeng},
      year={2022},
      eprint={2204.01691},
      archivePrefix={arXiv},
      primaryClass={cs.RO},
      url={https://arxiv.org/abs/2204.01691}, 
}

@inproceedings{zhang2024building,
    title     = {Building Cooperative Embodied Agents Modularly with Large Language Models},
    author    = {Zhang, Hongxin and Du, Weihua and Shan, Jiaming and Zhou, Qinhong and Du, Yilun and Tenenbaum, Joshua B. and Shu, Tianmin and Gan, Chuang},
    booktitle = {International Conference on Learning Representations (ICLR)},
    year      = {2024},
    url       = {https://proceedings.iclr.cc/paper_files/paper/2024/file/54b8b4e0b4ba4aad112e84f32e3b5dbb-Paper-Conference.pdf},
    note      = {arXiv preprint arXiv:2307.02485}
}

@article{sun2025collabovercooked,
    title   = {Collab-Overcooked: Benchmarking and Evaluating Large Language Models as Collaborative Agents},
    author  = {Sun, Zeyu and Zhang, Han and Wang, Cheng and Xu, Chenyang and Wang, Yizhou},
    journal = {arXiv preprint arXiv:2502.20073},
    year    = {2025},
    url     = {https://arxiv.org/abs/2502.20073}
}

@inproceedings{zhang2025cut,
  title={Cut the Crap: An Economical Communication Pipeline for {LLM}-based Multi-Agent Systems},
  author={Zhang, Guibin and Yue, Yanwei and Li, Zhixun and Yun, Sukwon and Wan, Guancheng and Wang, Kun and Cheng, Dawei and Yu, Jeffrey Xu and Chen, Tianlong},
  booktitle={Proceedings of the International Conference on Learning Representations (ICLR)},
  year={2025},
  url={https://openreview.net/forum?id=LkzuPorQ5L}
}

@misc{liu2025capo,
      title={CaPo: Cooperative Plan Optimization for Efficient Embodied Multi-Agent Cooperation}, 
      author={Jie Liu and Pan Zhou and Yingjun Du and Ah-Hwee Tan and Cees G. M. Snoek and Jan-Jakob Sonke and Efstratios Gavves},
      year={2025},
      eprint={2411.04679},
      archivePrefix={arXiv},
      primaryClass={cs.AI},
      url={https://arxiv.org/abs/2411.04679}, 
}

@misc{bruce2024genie,
    title         = {Genie: Generative Interactive Environments},
    author        = {Jake Bruce and Michael Dennis and Ashley Edwards and Jack Parker-Holder and Yuge Shi and Edward Hughes and Matthew Lai and Aditi Mavalankar and Richie Steigerwald and Chris Apps and Yusuf Aytar and Sarah Bechtle and Feryal Behbahani and Stephanie Chan and Nicolas Heess and Lucy Gonzalez and Simon Osindero and Sherjil Ozair and Scott Reed and Jingwei Zhang and Konrad Zolna and Jeff Dean and Nando de Freitas},
    year          = {2024},
    eprint        = {2402.15391},
    archivePrefix = {arXiv},
    primaryClass  = {cs.LG}
}

@misc{assran2025vjepa2,
  title         = {V-JEPA 2: Self-Supervised Video Models Enable Understanding, Prediction and Planning},
  author        = {Mido Assran and Adrien Bardes and David Fan and Quentin Garrido and Russell Howes and Mojtaba Komeili and Matthew J. Muckley and Ammar Rizvi and Claire Roberts and Koustuv Sinha and Artem Zholus and Sergio Arnaud and Abha Gejji and Ada Martin and Francois Robert Hogan and Daniel Dugas and Piotr Bojanowski and Vasil Khalidov and Patrick Labatut and Francisco Massa and Marc Szafraniec and Kapil Krishnakumar and Yong Li and Xiaodong Ma and Sarath Chandar and Franziska Meier and Yann LeCun},
  year          = {2025},
  eprint        = {2506.09985},
  archivePrefix = {arXiv},
  primaryClass  = {cs.CV}
}

@article{wang2023surveyagents,
  title={A Survey on Large Language Model based Autonomous Agents},
  author={Wang, Lei and Ma, Chen and Feng, Xueyang and Zhang, Zeyu and Yang, Hao and Zhang, Jingsen and Chen, Zhiyuan and Tang, Jiakai and Chen, Xu and Lin, Yankai and others},
  journal={Frontiers of Computer Science},
  volume={18},
  number={6},
  pages={186345},
  year={2024},
  publisher={Springer},
  doi={10.1007/s11704-024-40231-1}
}

@misc{ferrag2025reviewagents,
    title         = {From LLM Reasoning to Autonomous AI Agents: A Comprehensive Review},
    author        = {Mohamed Amine Ferrag and Norbert Tihanyi and Merouane Debbah},
    year          = {2025},
    eprint        = {2504.19678},
    archivePrefix = {arXiv},
    primaryClass  = {cs.AI},
    url           = {https://arxiv.org/abs/2504.19678}
}

@inproceedings{liu2025adatamp,
    title     = {{AdaTAMP}: Adaptive Task and Motion Planning with {LLM}-based Embodied Agents},
    author    = {Karan Baijal and Zhiwen Qiu and Jennifer Sun},
    booktitle = {ICRA Workshop on Language and Semantics of Task and Motion Planning},
    year      = {2025},
    url       = {https://dyalab.mines.edu/2025/icra-workshop/12.pdf}
}

@misc{du2025constrainedhumanaicooperationinclusive, 
    title         = {Constrained Human-AI Cooperation: An Inclusive Embodied Social Intelligence Challenge}, 
    author        = {Weihua Du and Qiushi Lyu and Jiaming Shan and Zhenting Qi and Hongxin Zhang and Sunli Chen and Andi Peng and Tianmin Shu and Kwonjoon Lee and Behzad Dariush and Chuang Gan}, 
    year          = {2025}, 
    eprint        = {2411.01796}, 
    archivePrefix = {arXiv}, 
    primaryClass  = {cs.AI}, 
    url           = {https://arxiv.org/abs/2411.01796}, 
}

@misc{mecattaf2025littleconversationlittleaction,
    title         = {A little less conversation, a little more action, please: Investigating the physical common-sense of LLMs in a 3D embodied environment}, 
    author        = {Matteo G. Mecattaf and Ben Slater and Marko Tešić and Jonathan Prunty and Konstantinos Voudouris and Lucy G. Cheke},
    year          = {2025},
    eprint        = {2410.23242},
    archivePrefix = {arXiv},
    primaryClass  = {cs.AI},
    url           = {https://arxiv.org/abs/2410.23242}, 
}

@misc{ramrakhya2025groundingmultimodalllmsembodied,
    title         = {Grounding Multimodal LLMs to Embodied Agents that Ask for Help with Reinforcement Learning}, 
    author        = {Ram Ramrakhya and Matthew Chang and Xavier Puig and Ruta Desai and Zsolt Kira and Roozbeh Mottaghi},
    year          = {2025},
    eprint        = {2504.00907},
    archivePrefix = {arXiv},
    primaryClass  = {cs.AI},
    url           = {https://arxiv.org/abs/2504.00907}, 
}

@misc{hsu2025groupthink,
    title         = {Group Think: Multiple Concurrent Reasoning Agents Collaborating at Token Level Granularity},
    author        = {Chan-Jan Hsu and Davide Buffelli and Jamie McGowan and Feng-Ting Liao and Yi-Chang Chen and Sattar Vakili and Da-shan Shiu},
    year          = {2025},
    eprint        = {2505.11107},
    archivePrefix = {arXiv},
    primaryClass  = {cs.CL},
    url           = {https://arxiv.org/abs/2505.11107}
}

@misc{ehsani2021manipulathorframeworkvisualobject,
    title         = {ManipulaTHOR: A Framework for Visual Object Manipulation}, 
    author        = {Kiana Ehsani and Winson Han and Alvaro Herrasti and Eli VanderBilt and Luca Weihs and Eric Kolve and Aniruddha Kembhavi and Roozbeh Mottaghi},
    year          = {2021},
    eprint        = {2104.11213},
    archivePrefix = {arXiv},
    primaryClass  = {cs.CV},
    url           = {https://arxiv.org/abs/2104.11213}, 
}

@misc{shridhar2020alfredbenchmarkinterpretinggrounded,
    title         = {ALFRED: A Benchmark for Interpreting Grounded Instructions for Everyday Tasks}, 
    author        = {Mohit Shridhar and Jesse Thomason and Daniel Gordon and Yonatan Bisk and Winson Han and Roozbeh Mottaghi and Luke Zettlemoyer and Dieter Fox},
    year          = {2020},
    eprint        = {1912.01734},
    archivePrefix = {arXiv},
    primaryClass  = {cs.CV},
    url           = {https://arxiv.org/abs/1912.01734}, 
}

@misc{ma2024doze,
    title         = {{DOZE}: A Dataset for Zero-Shot Object Navigation with Dynamic Obstacles},
    author        = {Ji Ma and Hongming Dai and Xinyu Sun and Zifan Wang and Yiqing Xu and Rong Xiong and Yue Wang},
    year          = {2024},
    eprint        = {2402.19007},
    archivePrefix = {arXiv},
    primaryClass  = {cs.RO},
    url           = {https://arxiv.org/abs/2402.19007}
}

@article{lee2025conversational,
    author  = {Soohwan Lee and Seoyeong Hwang and Dajung Kim and Kyungho Lee},
    title   = {Conversational Agents as Catalysts for Critical Thinking: Challenging Social Influence in Group Decision-making},
    year    = {2025},
    journal = {arXiv preprint arXiv:2503.14263},
    url     = {https://arxiv.org/html/2503.14263v1}
}

@misc{patania2025perspact,
    title         = {PerspAct: Enhancing LLM Situated Collaboration Skills through Perspective Taking and Active Vision}, 
    author        = {Sabrina Patania and Luca Annese and Anita Pellegrini and Silvia Serino and Anna Lambiase and Luca Pallonetto and Silvia Rossi and Simone Colombani and Tom Foulsham and Azzurra Ruggeri and Dimitri Ognibene},
    year          = {2025},
    eprint        = {2511.08098},
    archivePrefix = {arXiv},
    primaryClass  = {cs.RO},
    note          = {Accepted at IAS-19},
    url           = {https://arxiv.org/abs/2511.08098}
}

@article{camerer1989curse,
    title     = {The curse of knowledge in economic settings: An experimental analysis},
    author    = {Camerer, Colin and Loewenstein, George and Weber, Martin},
    journal   = {Journal of Political Economy},
    volume    = {97},
    number    = {5},
    pages     = {1232--1254},
    year      = {1989},
    publisher = {The University of Chicago Press}
}

@article{trott2025reevaluating,
  title={Re-evaluating Theory of Mind evaluation in large language models},
  author={Hu, Jennifer and Sosa, Felix and Ullman, Tomer},
  journal={Philosophical Transactions of the Royal Society B},
  volume={380},
  number={1932},
  year={2025},
  publisher={The Royal Society},
  doi={10.1098/rstb.2023.0499}
}

@inproceedings{li2025theoryofmind,
    title = "Theory of Mind for Multi-Agent Collaboration via Large Language Models",
    author = "Li, Huao  and
      Chong, Yu  and
      Stepputtis, Simon  and
      Campbell, Joseph  and
      Hughes, Dana  and
      Lewis, Charles  and
      Sycara, Katia",
    editor = "Bouamor, Houda  and
      Pino, Juan  and
      Bali, Kalika",
    booktitle = "Proceedings of the 2023 Conference on Empirical Methods in Natural Language Processing",
    month = dec,
    year = "2023",
    address = "Singapore",
    publisher = "Association for Computational Linguistics",
    url = "https://aclanthology.org/2023.emnlp-main.13/",
    doi = "10.18653/v1/2023.emnlp-main.13",
    pages = "180--192"
}

@misc{zhang2025bias,
      title={Understanding Bias Reinforcement in LLM Agents Debate}, 
      author={Jihwan Oh and Minchan Jeong and Jongwoo Ko and Se-Young Yun},
      year={2025},
      eprint={2503.16814},
      archivePrefix={arXiv},
      primaryClass={cs.LG},
      url={https://arxiv.org/abs/2503.16814}, 
}
\cleardoublepage
\appendix
\appendix

\section{Sample Interaction Logs}
\label{sec:appendix_logs}

To illustrate the "Push-Pull" dynamics described in Section \ref{sec:analysis}, we provide raw transcripts from the agent dyads. These logs highlight the contrast between successful active querying and failed passive obedience.

\subsection{Successful Episode: The "Pull" Mechanism}
In this episode, the Follower (Handicapped) actively queries the Leader when the initial instruction is ambiguous relative to its local view. This "Pull" forces the Leader to re-ground the instruction, leading to success.

\begin{table}[h!]
\small
\centering
\resizebox{\columnwidth}{!}{%
\begin{tabular}{|l|p{6.5cm}|}
\hline
\textbf{Agent} & \textbf{Message / Action} \\ \hline
\textbf{Leader} & \textit{Instruction:} "I see the Apple on the dining table. Move forward 3 steps." \\ \hline
\textbf{Follower} & \textit{Observation:} "I see a wall and a sofa. I do not see a dining table." \\ \hline
\textbf{Follower} & \textbf{\textit{Query (Pull):} "Which direction is the dining table from the sofa?"} \\ \hline
\textbf{Leader} & \textit{Reasoning:} "My previous instruction assumed global visibility. Correcting for sofa landmark." \\ \hline
\textbf{Leader} & \textit{Instruction:} "Turn 90 degrees right. The table is behind the sofa." \\ \hline
\textbf{Follower} & \textit{Action:} Rotates Right. \\ \hline
\textbf{Follower} & \textit{Observation:} "I see the table now." \\ \hline
\end{tabular}%
}
\caption{A successful "Pull" interaction. The Follower's query prevents a potential navigation error.}
\label{tab:success_log}
\end{table}

\subsection{Failed Episode: Privileged Information Bias}
In this failed episode, the Leader relies on its own egocentric frame of reference ("to your left"). The Follower fails to "Pull" (verify) and executes the action blindly, resulting in a collision.

\begin{table}[h!]
\small
\centering
\resizebox{\columnwidth}{!}{%
\begin{tabular}{|l|p{6.5cm}|}
\hline
\textbf{Agent} & \textbf{Message / Action} \\ \hline
\textbf{Leader} & \textit{State:} Target (Mug) is at $(x=1.5, z=3.0)$. \\ \hline
\textbf{Leader} & \textit{Instruction:} "The mug is just to your left. Move Left." \\ \hline
\textbf{Follower} & \textit{Observation:} "I see a cabinet to my left." \\ \hline
\textbf{Follower} & \textbf{\textit{Response (No Pull):} "Moving Left."} \\ \hline
\textbf{System} & \textit{Event:} Collision with Cabinet. \\ \hline
\textbf{Leader} & \textit{Instruction:} "You hit something. Move Left again." \\ \hline
\textbf{System} & \textit{Event:} Collision with Cabinet. (Episode Fail) \\ \hline
\end{tabular}%
}
\caption{A failed "Push" interaction. The Leader assumes its "Left" is the Follower's "Left," and the Follower fails to verify.}
\label{tab:failure_log}
\end{table}

\subsection{System Prompting (Context)}
The dyadic interaction is governed by a shared LLM core that alternates roles. The "Privileged Information Bias" is induced by explicitly injecting different state descriptions into the system prompt:
\begin{itemize}
    \item \textbf{Leader Context:} Receives full object list $O_{full}$ containing all objects in the scene with exact coordinates.
    \item \textbf{Follower Context:} Receives filtered object list $O_{partial}$ containing only objects within $2.0m$ and within a $90^{\circ}$ field of view.
\end{itemize}

\end{document}